\documentclass[journal]{IEEEtran}
\usepackage{algorithm,algpseudocode}
\usepackage{cuted}
\usepackage{tikz}
\usepackage[caption=false,font=footnotesize]{subfig}

\usepackage{cite}
\usepackage{amsmath,amssymb,amsfonts}
\usepackage{commath,bm}

\usepackage{url}
\usepackage{amsmath}
\DeclareMathOperator*{\argmax}{arg\,max}

\usepackage{array}
\usepackage{graphicx,wrapfig,lipsum}
\usepackage{mathtools}
\usepackage{multirow}

\ifCLASSINFOpdf
\else
\fi
\begin{document}
%
\title{Sensor Selection via GFlowNets: A Deep Generative Modeling Framework to Navigate Combinatorial Complexity}
%
%
%

\author{Spilios~Evmorfos,
        Zhaoyi~Xu,
        and~Athina~Petropulu,
\thanks{The authors are with the ECE Department at Rutgers, the State University of New Jersey, New Brunswick, NJ
\newline Work supported by ARO under Grants W911NF2110071 and W911NF2320103 and by NSF under Grant ECCS-2335876
}
}

\maketitle

\begin{abstract}
The performance of sensor arrays in sensing and wireless communications improves with more elements, but this comes at the cost of increased energy consumption and hardware expense. This work addresses the challenge of selecting $k$ sensor elements from a set of $m$ to optimize a generic Quality-of-Service metric. Evaluating all $\binom{m}{k}$ possible sensor subsets is impractical, leading to prior solutions using convex relaxations, greedy algorithms, and supervised learning approaches. The current paper proposes a new framework that employs deep generative modeling, treating sensor selection as a deterministic Markov Decision Process where sensor subsets of size $k$ arise as terminal states. Generative Flow Networks (GFlowNets) are employed to model an action distribution conditioned on the state. Sampling actions from the aforementioned distribution ensures that the probability of arriving at a terminal state is proportional to the performance of the corresponding subset. Applied to a standard sensor selection scenario, the developed approach outperforms popular methods which are based on convex optimization and greedy algorithms. Finally, a multiobjective formulation of the proposed approach is adopted and applied on the sparse antenna array design for Integrated Sensing and Communication (ISAC) systems. The multiobjective variation is shown to perform well in managing the trade-off between radar and communication performance.
\end{abstract}


\IEEEpeerreviewmaketitle

\section{Introduction}
In a sensor array, the measurement captured by each sensor is a function of the quantity of interest. The objective is to accurately determine that quantity from the diverse set of measurements. Generally, a larger number of measurements leads to more accurate estimates of the quantity of interest. In cases like MIMO radar \cite{li2008mimo}, a larger array leads to enhanced sensing resolution \cite{4176505}, reflecting overall improved performance with more elements. However, operating a large number of sensor elements escalates operational costs (power consumption, monetary expenses, etc.). To address this, a common strategy is to activate only a subset of deployed sensor elements, aiming to optimize a specific performance objective. Given $m$ sensor elements and a desire to activate only $k < m$ of them, the number of possible subsets is $\binom{m}{k}$. When both $m$ and $k$ are substantial, exhaustive search becomes impractical. 
As a result, numerous research endeavors have concentrated on devising computationally efficient methods to choose the optimal subset, addressing the hurdles presented by the combinatorial explosion within the solution space.

\textbf{Convex Optimization}: Given that the original selection problem entails integer constraints, certain approaches utilize convex relaxations to create surrogate problems that are convex in continuous domains. They employ convex optimization techniques, which have demonstrated efficacy in addressing such problems. Subsequently, the solutions of the convex surrogates are transformed into feasible solutions corresponding to the selection of a specific sensor subset.
The approach outlined in \cite{4663892} selects a sensor subset aimed at minimizing the estimation error of a vector of interest. Each sensor measurement corresponds to a linear function of the vector of interest. The  methodology in \cite{4663892} involves a convex relaxation followed by the application of an Interior Point Method (IPM) \cite{gondzio2012interior} approach. A similar method is proposed in \cite{6760894}.  The work by \cite{wielgus2024general} examines the problem of sensor selection in the context of broadband receiver beamforming, where the goal is to choose the subset that minimizes the $L_{2}$ distance between the resulting beampattern and a desired one and the solution is obtained via a convex relaxation and a Branch-and-Bound approach \cite{boyd2007branch}. The work of  \cite{nosrati2017receiver} introduces a semidefinite programming method to choose a transmitter-receiver antenna pair that maximizes the separation between desired and undesired directions of arrival. The authors in \cite{richard1991design} demonstrate that designing a sparse beamforming array with unequally spaced antennas, achieved by selecting specific elements from a uniform array, can be formulated as a quasi-norm minimization problem. Additionally, they introduce a simplex search method as a proposed solution.

\textbf{Greedy Methods}: Greedy approaches are characterized by their sequential and
iterative selection of sensors for the final subset. These methods often leverage the specific structure of the optimization objective to devise a near-optimal greedy selection strategy, subject to certain assumptions.
A greedy approach is presented in \cite{6031934} to select a subset that minimizes the Cramer-Rao Bound of target estimates within the context of distributed multiple-radar systems. The authors in \cite{5717225} examine the same setting as the work in \cite{4176505} and formulate the sensor selection problem as the maximization of a monotone submodular function over uniform matroids. They develop an efficient greedy selection approach that yields a near-optimal solution to the selection problem. Similar analysis and approach are provided in \cite{majumder2023greedy}, albeit in the context of heterogeneous sensors, where sensors exhibit varying precision levels and operational costs.

\textbf{Supervised Learning}: Supervised machine learning approaches frame the sensor selection problem as a multilabel classification task, training a function approximator on a dataset of annotated examples.
In the work of \cite{joung2016machine}, Support Vector Machines \cite{hearst1998support} are proposed for sparse antenna array design in wireless communications. Similarly, in \cite{elbir2019cognitive}, the antenna selection problem is reformulated as a classification problem, and a Convolutional Neural Network \cite{lecun1995convolutional} serves as the classifier. Meanwhile, \cite{vu2021machine} introduces a supervised deep learning method for learning optimal antenna selection and precoding matrices.
For massive MIMO systems, \cite{lin2021deep} proposes a supervised deep learning approach to antenna selection based on a channel state information extrapolation metric. In the context of MIMO transmit beamforming, \cite{diamantaras2021sparse} advocates a supervised learning approach utilizing a neural network composed of elementary operations resembling self-attention \cite{vaswani2017attention}. A similar strategy is employed in \cite{xu2022cramer} for sensor selection in Integrated Sensing and Communication (ISAC) systems.

\textbf{Pitfalls of existing approaches}: Analytical approaches, which include both convex optimization methods and greedy selection strategies, have demonstrated notable success in various sensor selection scenarios. However, they have certain drawbacks. Firstly, they are susceptible to getting stuck at local optima. Additionally, they are criterion-dependent, relying on specific properties of the optimization objective to facilitate the application of convex relaxations or greedy selection. This dependency on the optimization criterion poses challenges in sensor selection applications when the objective does not align with the assumptions inherent in these analytical methods. For example, the sensor selection objective discussed in \cite{xu2022cramer}, tailored for ISAC systems, does not adhere to the submodular function property, making the development of a greedy approach difficult. Moreover, the application of convex relaxations and approximations for employing off-the-shelf convex optimization algorithms is not straightforward.
On a different note, supervised learning methods demand a significant amount of annotated data, a resource notably scarce within the pertinent domains of interest. 

\textbf{Contributions}: The current paper introduces a novel framework for addressing sensor selection problems, aiming to overcome the limitations associated with existing paradigms. This proposed framework is distinctive in its unsupervised nature, eliminating the reliance on annotated data, and it is criterion-agnostic, thus removing the necessity for specific assumptions regarding the selection objective (e.g., convexity, differentiability, submodularity, etc.).

The innovation of the proposed method lies in its modeling of the selection process as a deterministic Markov Decision Process (MDP) with a single root. Each sensor subarray selection of diserable size corresponds to a root-to-terminal-node trajectory in the MDP. Non-terminal states are assigned a zero reward, while the reward for each terminal node is determined by evaluating the objective for the corresponding subset. This formulation sets the goal of learning a distribution over actions, conditioned on the state, ensuring the cumulative probability of reaching a terminal state is proportional to the state's reward (the performance of the corresponding subset).

The distribution over actions is parametrized by the Generative Flow Networks (GFlowNets) paradigm. GFlowNets introduce a flow quantity to amortize the learning cost of distributions over composite objects like MDPs. This approach transforms flow-matching equations into a learning objective, shifting the distribution learning problem from the combinatorial action space to the continuous parameter space of a function approximator. Stochastic gradient descent methods, recognized for their generalization performance to unseen examples, can be leveraged to adjust the parameters of the function approximator.

Furthermore, the GFlowNet paradigm is applied to the sensor selection setting examined in previous works \cite{4663892, 5717225}, showcasing superior performance compared to the convex optimization approach of \cite{4663892} and the greedy approach of \cite{5717225}. Specifically, our approach was implemented for selecting $15$ sensors from a pool of $100$. With approximately $2 \times 10^{17}$ possible subsets within the feasible set, we employ a small neural network featuring $3$ layers with $150$ neurons each layer, trained over $40,000$ root-to-leaf trajectories. This signifies that out of the $2 \times 10^{17}$ subsets, we utilize at most $40,000$ (which corresponds to less than $5 \times 10^{-10}\%$ of the possible ones) for training. Furthermore, the GFlowNet paradigm for sensor selection is modified to accommodate a multiobjective setting for sensor selection and is applied to the ISAC sensor selection scenario that is outlined in \cite{xu2022cramer}.

The contributions of the current work are outlined below:
\begin{itemize}
    \item The current paper introduces an innovative approach to sensor selection problems that is both unsupervised and criterion-agnostic, allowing for a broad applicability across different settings without the need for specific assumptions and annotated data.
    \item The proposed approach models the sensor selection process as a deterministic MDP with a singular root and sensor subsets of desirable size arising as terminal states.
    \item The GFlowNet paradigm is employed to parametrize the distribution over actions in the MDP. It is trained such that the cumulative probability of arriving at a terminal state is proportional to the performance of the corresponding subset.
    \item The proposed approach is applied on a sensor selection setting for linear estimation and showcased to surpass the performance of a convex optimization method and a greedy selection approach.
    \item The paper extends the proposed approach to a multiobjective formulation and showcases its versatility to a sparse array design problem for ISAC systems.
\end{itemize}

\underline{\textit{Notation}}: We denote the matrices and vectors by bold uppercase and bold lowercase letters, respectively.  The operators $\left(\cdot\right)^{\mathsf{T}}$ and $\left(\cdot\right)^{H}$ denote transposition and conjugate transposition respectively. Caligraphic letters will be used to denote sets. The $\ell_{p}$-norm of
$\boldsymbol{x}\in\mathbb{R}^{n}$ is $\left\Vert \boldsymbol{x}\right\Vert _{p}\triangleq\left(\sum_{i=1}^{n}\left|x\left(i\right)\right|^{p}\right)^{1/p}$,
for all $\mathbb{N}\ni p\ge1$. The expectation of a random vector $\mathbf{x}$ is denoted as $\mathbb{E}\left( \mathbf{x}\right)$. Continuous sets are denoted by $[ \cdot]$ while discrete sets are denoted by $\{ \cdot\}$.

\section{Generative Flow Networks}
This section offers an overview of the GFlowNet  framework \cite{bengio2023gflownet}. Consider a deterministic MDP with $\mathcal{S}$ as the set of states and $\mathcal{X} \subset \mathcal{S}$ as the set of terminal states. Let $\mathcal{A}$ represent the set of discrete actions, and $\mathcal{A}(\mathbf{s})$ denote the set of permissible actions at state $\mathbf{s}$. The MDP is represented as a Directed Acyclic Graph (DAG).  All leaf nodes possess positive rewards, all intermediate states bear zero reward ($\mathcal{R}(\mathbf{s}) = 0 \hspace{0.1cm} \forall \mathbf{s} \notin \mathcal{X}$), and a unique root $\mathbf{s}_{0}$ exists. The DAG is non-injective, implying that different action sequences (starting from the root) can lead to the same state. The primary objective is to learn an action selection policy such that the probability of reaching a terminal state is proportional to the terminal state's reward.

The GFlowNet conceptualizes the MDP as a flow network, where the flow originates from the root, and each leaf node serves as a flow sink. Assuming that action $\mathbf{a}$ is executed at state $\mathbf{s}$, the subsequent state $\mathbf{s}'$ is denoted as $T(\mathbf{s},\mathbf{a}) = \mathbf{s}'$. Given the deterministic nature of the MDP, $T(\mathbf{s},\mathbf{a})$ is uniquely determined for each pair. The flow along edge $(\mathbf{s}, \mathbf{a})$ is represented as $F(\mathbf{s},\mathbf{a})$, and the total flow passing through state $\mathbf{s}$ is denoted as $F(\mathbf{s})$. To adhere to flow balance conditions, the incoming flow of each state must equal the outgoing flow. For any node $\mathbf{s}'$, the in-flow is:
\begin{equation}
    F(\mathbf{s}') = \mkern-24mu\sum_{\mathbf{s},\mathbf{a}: T(\mathbf{s},\mathbf{a}) = \mathbf{s}'} \mkern-24mu F(\mathbf{s},\mathbf{a}).
\end{equation}
On the other hand, the out-flow can be defined as:
\begin{equation}
    F(\mathbf{s}') = \mkern-12mu\sum_{\mathbf{a}' \in \mathcal{A}(\mathbf{s}')} \mkern-12mu F(\mathbf{s}', \mathbf{a}').
\end{equation}
The flow of each state is the sum of the reward of the state and the flow of all outgoing edges:
\begin{equation}
    F(\mathbf{s}) = R(\mathbf{s}) + \sum_{\mathbf{a} \in \mathcal{A}(\mathbf{s})} F(\mathbf{s},\mathbf{a}).
\end{equation}
This implies that the flow of each terminal state is the reward of the state:
\begin{equation}
    F(\mathbf{x}) = R(\mathbf{x}) > 0.
\end{equation}
Since the flow of each state is also equal to the flow of all incoming edges, the flow-matching equation that holds on each node $\mathbf{s}^{'}$ is:
\begin{equation}
\sum_{\mathbf{s},\mathbf{a}: T(\mathbf{s},\mathbf{a}) = \mathbf{s}'} F(\mathbf{s},\mathbf{a}) = R(\mathbf{s}') + \sum_{\mathbf{a}' \in \mathcal{A}(\mathbf{s}')} F(\mathbf{s}', \mathbf{a}').
\end{equation}

Under the assumption that the flow of each state $F(\mathbf{s})$ and the flow of each edge $F(\mathbf{s}, \mathbf{a})$ are known and satisfy the above equation, if actions are sampled on each state (starting from the root) based on the ratio
\begin{equation}
    \pi(\mathbf{a} | \mathbf{s}) = \frac{F(\mathbf{s},\mathbf{a})}{F(\mathbf{s})},
    \label{sampling policy}
\end{equation}
the following results hold \cite{bengio2021flow}:
\begin{enumerate}
    \item The flow of the root (also known as the partition function of the DAG) is equal to the sum of all the rewards of the terminal nodes:
    \begin{equation}
        F(\mathbf{s}_{0}) = \sum_{\mathbf{x} \in \mathcal{X}} R(\mathbf{x}).
    \end{equation}
    \item The probability of reaching a terminal state $\mathbf{x}$ is the ratio of the reward of the state over the partition function:
    \begin{equation}
        \pi(\mathbf{x}) = \frac{R(\mathbf{x})}{\sum_{\mathbf{x}' \in \mathcal{X}} R(\mathbf{x}')} = \frac{R(\mathbf{x})}{F(\mathbf{s}_{0})}.
    \end{equation}.
\end{enumerate}

The GFlowNet paradigm involves parameterizing the flow $F$ using a function approximator $F_{\mathbf{w}}$ selected from a class with sufficient expressivity, such as a deep neural network. Trajectories of the MDP are sampled, and for each state $\mathbf{s}'$, the following objective, referred to as flow-matching objective, is minimized via gradient descent on $\mathbf{w}$:
\begin{equation}
\begin{aligned}[b]
    L_{\mathbf{w}} ( \mathbf{s}') = \mkern-24mu\sum_{\mathbf{s},\mathbf{a}:T(\mathbf{s},\mathbf{a}) = \mathbf{s}'} \mkern-24mu F_{\mathbf{w}}(\mathbf{s},\mathbf{a}) - R(\mathbf{s}') - \mkern-18mu\sum_{\mathbf{a}' \in \mathcal{A}(\mathbf{s}')} \mkern-15mu F_{\mathbf{w}}(\mathbf{s}',\mathbf{a}').
    \end{aligned} \label{flow matching}
\end{equation}

\section{Sensor Selection via GFlowNets}
\label{general method}
The problem of sensor selection entails the task of choosing a subset of $k$ active sensors from a sensor array containing $m$ elements to optimize a performance metric $Q$. Given the $\binom{m}{k}$ possible subsets to consider, the primary challenge involves identifying the subset that maximizes (or minimizes) the performance metric.

To effectively leverage the GFlowNet paradigm, let us transform the sensor selection problem into the task of sampling terminal states from a MDP, which is defined as follows: The state space is discrete, with each state $\mathbf{s}$ is represented as a binary vector of $m$ elements. Within this vector, ``1" elements denote active sensor positions, while ``0" elements indicate inactive sensors. The initial state, denoted as $\mathbf{s}_{0}$, is the zero vector where all elements are inactive. The action space is also discrete, allowing actions at each state to involve adding one extra active element from the set of inactive ones.

Terminal states in the MDP are characterized by state vectors with exactly $k$ active elements ($\mathbf{s} \equiv \mathbf{x} \in \mathcal{X}$ iff $\lVert \mathbf{s}\rVert_{0} = k$). Intermediate states, where $\lVert \mathbf{s} \rVert_{0} < k$, yield zero rewards.

\begin{figure}[ht] 
\centering
 \vspace{-1mm}
{\label{fig:lossideal}\includegraphics[width = 7.5cm]{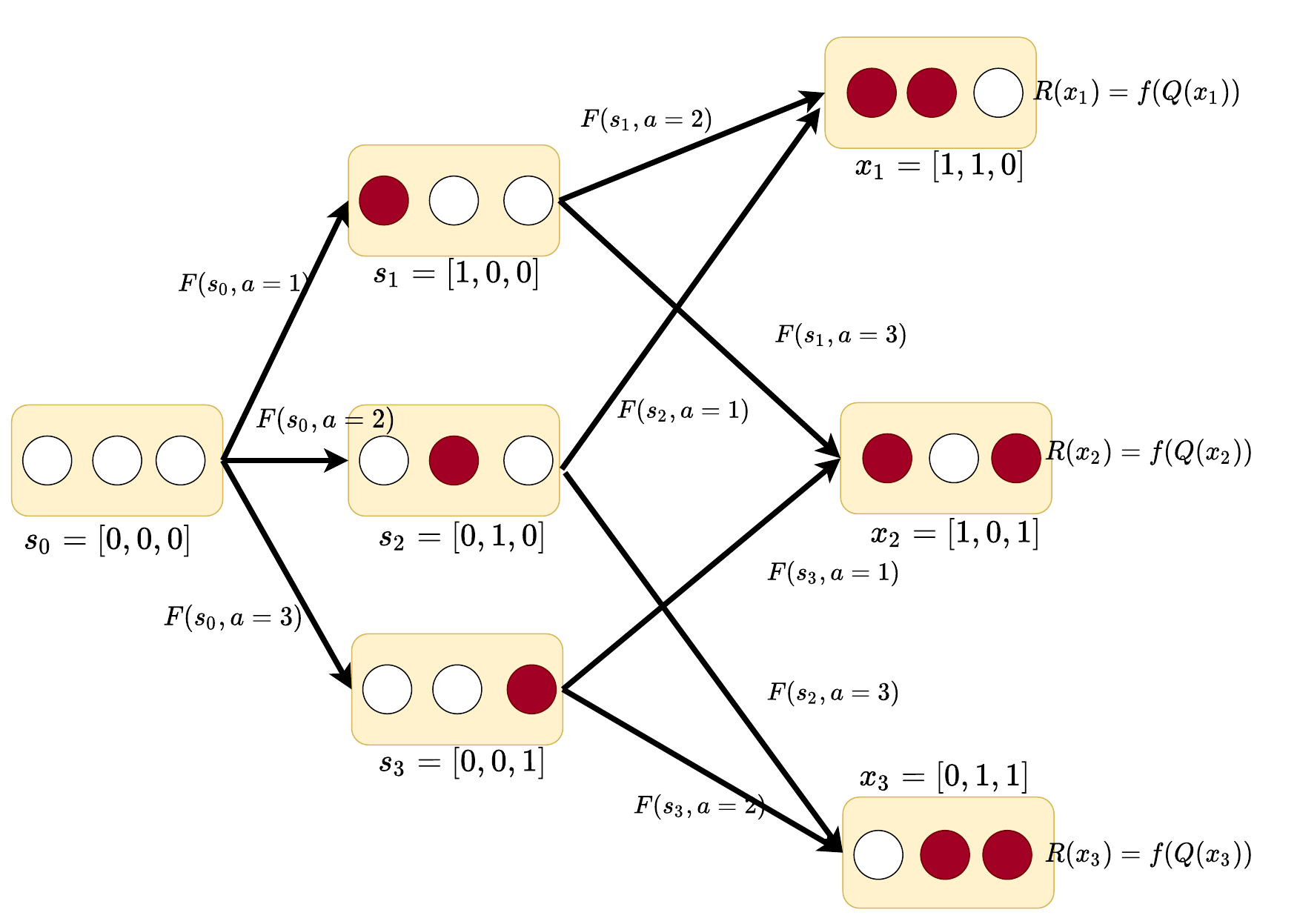}}%
\caption{\small The Sensor Selection MDP pertains to the task of selecting 2 sensor elements from a total of 3. In the visual representation, the red circles signify active elements, while the white circles indicate inactive elements. } \label{MDP flow}
\vspace{-2mm}
\end{figure}

For all terminal states $\mathbf{x}$, the reward is determined based on the evaluation of a function of the performance objective, represented as $f(Q(\mathbf{x}))$. The corresponding general graph for the sensor selection MDP is visually depicted in Fig. \ref{MDP flow}. The choice of the function of the performance metric is application-dependent. In the simplest case, assuming $Q(\mathbf{x}) > 0 \hspace{0.1cm} \forall \mathbf{x}$, the reward function is defined as $R(\mathbf{x}) = Q(\mathbf{x})$ for maximization goals, or as $R(\mathbf{x}) = \frac{1}{Q(\mathbf{x})}$ for minimization objectives.

The algorithm proposed for sensor selection, identified as GFlowNet for Sensor Selection (\textbf{GFLOW-SS}), is outlined in Algorithm \ref{gflow algo}. \textbf{GFLOW-SS} parameterizes the MDP flow using a neural network, and trajectories are sampled on-policy. Stochastic gradient descent is employed to update the network's weights based on the flow-matching loss (eq. \eqref{flow matching}). Following training, subsets can be sampled as terminal states utilizing the flow-induced action sampling policy defined in eq. \eqref{sampling policy}.

Two important aspects should be highlighted regarding the proposed approach. First, the training is unsupervised, eliminating the need for offline data preprocessing and annotation. This sets the proposed approach apart from previous machine learning methods addressing similar problems \cite{joung2016machine, elbir2019cognitive, lin2021deep}.
Secondly, a notable feature of this method is its objective-agnostic nature, making it independent of annotations and free from assumptions regarding the smoothness, differentiability, or convexity of the system's performance objective. The only requirement is the capability to evaluate the objective for a given subset of sensor elements. This contrasts with analytical methods, as exemplified in \cite{6031934, 4663892, 5717225, nosrati2017receiver, 9676708}, which rely on application-specific properties of the objective for employing convex relaxations or greedy selection.

{ 
\begin{algorithm}
\caption{\textbf{GFLOW - SS}}
\label{gflow algo}
\begin{algorithmic}
\State{Initialize $F_{\mathbf{w}}$, $\zeta \in [0,1]$ for exploration,  learning rate $\eta$}
\For {\texttt{all root-to-leaf trajectories}}
    \State{$\mathbf{s} = \mathbf{s}_{0} = [0]^{m}$}
    \For{$k-1$ transitions}
        \State {Sample $z \sim \mathcal{U}(0,1)$ (Uniform distribution)}
        \State {If $z < \zeta$ choose $\mathbf{a} \in \mathcal{A}(\mathbf{s})$ randomly}
        \State {If $z >= \zeta$ choose $\mathbf{a} = \argmax_{\mathbf{a}'} F_{\mathbf{w}}(\mathbf{s}, \mathbf{a}')$}
        \State{Apply action $\mathbf{a}$, compute sub. state $\mathbf{s}'$}
        \State {$\mathbf{w}' \rightarrow \mathbf{w} - \eta \nabla_{\mathbf{w}} L_{\mathbf{w}}(\mathbf{s}') $ \hspace{0.1cm} eq. \eqref{flow matching} \hspace{0.1cm} $R(\mathbf{s}') = 0$}
        \State { $\mathbf{s}$ = $\mathbf{s}'$}
    \EndFor
        \State {Sample $z \sim \mathcal{U}(0,1)$}
        \State {If $z > \zeta$ choose $\mathbf{a} \in \mathcal{A}(\mathbf{s})$ randomly}
        \State {If $z \leq \zeta$ choose $\mathbf{a} = \argmax_{\mathbf{a}'} F_{\mathbf{w}}(\mathbf{s}, \mathbf{a}')$}
        \State{Apply action $\mathbf{a}$, compute terminal state $\mathbf{x}$}
        \State {$\mathbf{w}' \rightarrow \mathbf{w} - \eta \nabla_{\mathbf{w}} L_{\mathbf{w}}(\mathbf{x}) $ \hspace{0.02cm} $R(\mathbf{x}) = Q(\mathbf{x}) \text{ or } \frac{1}{Q(\mathbf{x})}$}
\EndFor
\end{algorithmic}
\end{algorithm}
}

\section{Sensor Selection for Linear Estimation}
The method delineated in Section \ref{general method} is inherently versatile. In this section, we tailor it specifically to address the sensor selection problem, previously tackled through convex optimization in \cite{4663892} and through greedy selection in \cite{5717225}. Subsequently, we evaluate its performance in direct comparison to the two existing methodologies.

Consider $m$ linear sensor measurements  $y_{i}$ corresponding to the vector of interest $\mathbf{z} \in \mathbb{R}^{n}$.  Each measurement is corrupted by independent noise:
\begin{equation}
    y_{i} = \mathbf{a}_{i}^{\mathsf{T}} \mathbf{z} + v_{i}, \hspace{0.3cm} i=1,\dots, m
\end{equation}
The noise scalars $v_{1},\dots,v_{m}$ are independent identically distributed $\mathcal{N}(0, \sigma^{2})$ random variables. The measurement vectors $\mathbf{a}_{1},\dots,\mathbf{a}_{m}$ span $\mathbb{R}^{n}$. 




The sensor selection problem is to choose $k < m$ out of $m$ sensor vectors $\mathbf{a}_{i}$ such that the log of the determinant of the sum of outer products is maximized:
\begin{equation}
\begin{aligned}
& \text{maximize}_{\mathcal{S}}
& & \log \text{det} \left( \sum_{i\in\mathcal{S}} \mathbf{a}_{i} \mathbf{a}_{i}^{\mathsf{T}}\right) \\
& \text{subject to}
& & \| \mathcal{S} \| = k\\
\end{aligned}
\label{original problem}
\end{equation}
where $\mathcal{S} \subseteq \{1,\dots, m \}$ and $\| \mathcal{S} \|$ denotes the cardinality of set $\mathcal{S}$. The solution of the above optimization problem minimizes the estimation error of the maximum likelihood estimate of $\mathbf{z}$ \cite{5717225}.

\subsection{Convex Optimization}
The above problem is categorized as Boolean-convex, since the objective function is convex and the constraints are Boolean \cite{boyd2004convex}. In the work of \cite{4663892}, it is proposed to relax the boolean constraints into convex constraints, thereby facilitating the solution of the problem through convex optimization techniques. Specifically, the authors introduce a variable $\mathbf{x} \in \mathbb{R}^{m}$ and reformulate the original optimization problem into the relaxed version presented below:
\begin{equation}
\begin{aligned}
& \text{maximize}_{\mathbf{x}}
& & \log \text{det} \left( \sum_{i=1}^{m} x_{i}\mathbf{a}_{i} \mathbf{a}_{i}^{\mathsf{T}}\right) \\
& \text{subject to}
& & \| \mathbf{x} \|_{1} = k\\
& & &  x_{i} \in [0,1] \hspace{0.2cm} i=1,\dots, m\\
\end{aligned}
\end{equation}
The above problem can be effectively addressed using  IPMs \cite{gondzio2012interior}. The solution derived from solving the relaxed problem serves as an upper bound for the optimal solution of the original Boolean-convex problem \cite{4663892}. To obtain a solution for the initial problem, one can derive it from the solution of the relaxed version by arranging the elements in descending order, selecting the first $k$ elements (those with the highest magnitude), and setting their corresponding indices to $1$, while the rest are set to $0$.

We denote this approach as \textbf{CVX-OPT-SS} throughout the manuscript. While there is no established theoretical result defining the optimality gap between the solution of the relaxed problem and the feasible (binary) solution derived from it, it is widely acknowledged that the binary solution is generally suboptimal. The authors of \cite{4663892} advocate for the use of local search methods (\cite{wynn1972results, fedorov2013theory}) to refine the binary solution of the IPM, with such local methods demonstrating an additional increase in performance. In the experiments conducted in this paper, we intentionally omit these local approaches as they are orthogonal to the convex optimization approach and can be independently applied on top of the solution of any approach (including the subsequent greedy approach and the proposed \textbf{GFLOW-SS}).

\subsection{Greedy Selection}

In \cite{5717225}, the  optimization problem is posed as follows:

\begin{equation}
\begin{aligned}
& \text{maximize}_{\mathbf{x}}
& & \log \text{det} \left( \sum_{i=1}^{m} x_{i}\mathbf{a}_{i} \mathbf{a}_{i}^{\mathsf{T}} + \epsilon \mathbf{I} \right) \\
& \text{subject to}
& & \| \mathbf{x} \|_{1} = k\\
& & &  \mathbf{x}_{i} \in \{0,1\} \hspace{0.2cm} i=1,\dots, m\\
\end{aligned}
\end{equation}
where $\mathbf{I}$ denotes the identity matrix and $\epsilon$ is a small positive constant. The authors show that the objective function is a monotone submodular function, and the feasible set constitutes a uniform matroid \cite{calinescu2011maximizing}. Building upon this, \cite{5717225} introduces a straightforward greedy algorithm, commencing with a zero vector $\mathbf{x}_{0}$ and incrementally adding ``1" elements in accordance with a greedy criterion. Throughout the manuscript, we refer to this approach as \textbf{GREEDY-SS}.

\subsection{GFlowNet Approach}
Here we adapt the proposed approach, outlined in Section \ref{general method}, to the original problem formulation of eq. \eqref{original problem}. This involves defining the \underline{reward function} for each terminal node (corresponding to the performance of the respective subset) and determining the \underline{parameterization} of the flow network.

Given that the GFlowNet framework is akin to the amortization of an energy function to a sampling distribution \cite{bengio2021flow}, it is essential for the reward of each terminal node to be nonnegative. The goal is to identify the subset-terminal node that maximizes the resulting determinant of the sum of outer products. As the determinant of a matrix falls within the range $[-\infty, \infty]$, the sigmoid function, being strictly increasing, is employed to compress the output into the range $[0,1]$. The sigmoid function is defined as follows:

\begin{equation*}
    \sigma(x) = \frac{1}{1 + e^{-x}}
\end{equation*}
If we assume that a terminal node $\mathbf{x} \in \mathcal{X}$ corresponds to a determinant $D(\mathbf{x}) = \text{det}\left( \sum_{i=1}^{m} x_{i} \mathbf{a}_{i}\mathbf{a}_{i}^{\mathsf{T}}\right)$, the reward assigned to node $\mathbf{x}$ is the sigmoid of the determinant $D(\mathbf{x})$, i.e.,
\begin{equation}
    R(\mathbf{x}) = \sigma\left( D(\mathbf{x})\right).
\end{equation}
As the reward signal is the sole learning signal during updates and influences the estimation of the flow for each node through bootstrapping, we opt to scale the output of the sigmoid by a large constant $c$. This multiplication transforms the range of the reward signal into $[0, c]$:
\begin{equation}
    R(\mathbf{x}) = \sigma\left( D(\mathbf{x})\right) c.
    \label{reward scaling}
\end{equation}
Regarding the parametrization of the flow, a straightforward option is to use a multilayer perceptron neural network, where each layer comprises an affine transformation and Rectified Linear (ReLU) activation. However, recent findings in deep learning for function approximation \cite{tancik2020fourier} suggest that multilayer perceptrons with ReLU activations struggle to capture high-frequency components of a target signal, particularly when dealing with binary inputs. Hence, we adopt an alternative approach proposed in \cite{9676432,10285914, tancik2020fourier, mildenhall2021nerf}. In this approach, the initial layer of the perceptron is replaced with a learnable Fourier features kernel, consisting of an affine transformation followed by a sinusoidal activation function. The elements of the Fourier kernel are initialized from a zero-mean Gaussian distribution, with the variance treated as a tunable hyperparameter.

\subsection{Experiments}
In this subsection, we perform a series of experiments to compare the two previously established methods (\textbf{CVX-OPT-SS} and \textbf{GREEDY-SS}) with our proposed approach, which leverages the GFlowNet paradigm (\textbf{GFLOW-SS}).

We consider an instance of the setting, where $m=100$ and the number of selected sensors ranges from $5$ to $15$. The projection (measurement) vectors $\mathbf{a}_{1},\dots, \mathbf{a}_{m}$ are sampled independently, in a random fashion, from a Gaussian distribution $\mathcal{N}\left(0,1\right)$.

Concerning the \textbf{GREEDY-SS} approach, we set the $\epsilon$ parameter to $1$e-$12$. This parameter is essential for ensuring that the objective function remains monotone submodular and is expected to be kept small.

\begin{figure}[ht] 
\centering
 \vspace{-1mm}
{\label{fig:lossideal}\includegraphics[width = 9.0cm]{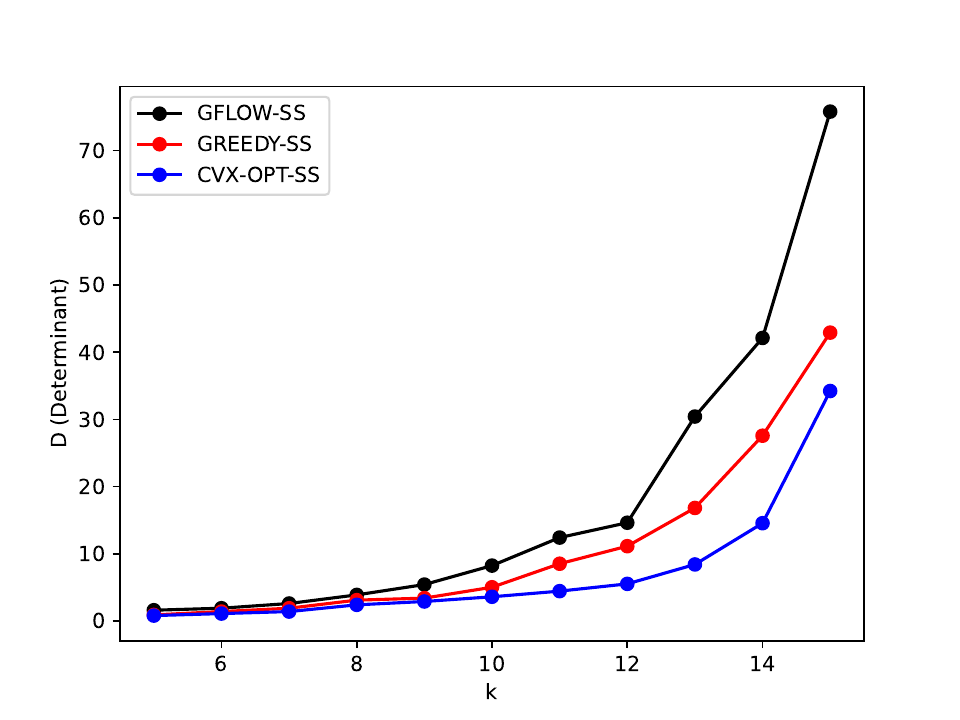}}%
\vspace{-0.5cm}
\caption{\small Comparison of the performance of \textbf{GFLOW-SS}, \textbf{GREEDY-SS} and \textbf{CVX-OPT-SS} for the problem of selecting $k$ sensors out of $100$. The parameter $k$ ranges from $5$ to $15$. Each point corresponds to the average performance of the respective approach over $8$ different instantiations. The \textbf{GFLOW-SS} approach is trained for $40000$ root-to-leaf MDP trajectories for every instantiation of the problem. This corresponds to $40000 \times k$ gradient descent steps.} \label{compare linear estimation}
\vspace{-2mm}
\end{figure}

Concerning the proposed \textbf{GFLOW-SS} approach, we opt for a $3$-layer dense neural network as the parametrization for the flow. Each layer is configured with $150$ neurons. The activation function for the second layer is ReLU, and the first layer corresponds to the Fourier kernel. The elements of this kernel are i.i.d samples from a zero-mean Gaussian distribution with a standard deviation of $0.1$ at initialization. The reward scaling parameter $c$ (eq. \ref{reward scaling}) is chosen to be $1000$. The Adam optimizer \cite{kingma2014adam} is used as the stochastic optimization algorithm with a learning rate of $2$e-4. 

The results are illustrated in Fig. \ref{compare linear estimation}. Each dot in the plot represents the performance of a solution (a $k$-size subset) generated by the respective approach. For each value of $k$ and each algorithm, the problem is solved from scratch. Each plotted point reflects the average over $8$ different instantiations of the problem.
For every value of $k$, the \textbf{GFLOW-SS} method undergoes training with a total of $40000$ root-to-leaf trajectories of the MDP. The training involves approximately $40000 \times k$ gradient steps. Consequently, the \textbf{GFLOW-SS} method explores a maximum of $40000$ different subsets.
It's crucial to note that for the smallest experiment ($\binom{100}{5}$), the number of potential subsets is approximately $75$ million. In contrast, for the largest experiment ($\binom{100}{15}$), the cardinality of the solution set escalates to $2 \times 10^{17}$. Clearly, $40000$ subsets represent only a minute fraction of the possible solutions in each case.
The \textbf{GFLOW-SS} method consistently outperforms both previous methods, occasionally yielding subsets with determinants up to twice as large.

The trained GFlowNet explicitly models an action sampling distribution conditioned on the state of the MDP of interest. Assuming that the training process has identified a point in the flow approximator parameter space satisfying the flow-matching objective, sampling actions from the corresponding distribution ensures that each terminal state of the MDP is sampled with a probability proportional to its reward. In the context of sensor selection, each k-size subset is sampled with a probability proportional to the corresponding system determinant (after being transformed by the sigmoid function and scaled). To sample the best subset from the trained GFlowNet, one starts from the MDP root (the state represented as all ``0"s) and greedily selects the action that maximizes the estimated flow at each state. The performance of \textbf{GFLOW-SS}, depicted in Fig. \ref{compare linear estimation}, corresponds to the subset sampled in that greedy fashion.

\subsubsection{Diversity in Generation}
In contrast to \textbf{GREEDY-SS} and \textbf{CVX-OPT-SS}, which offer a method to identify only the best possible subset, the \textbf{GFLOW-SS} approach implicitly establishes a ranking among the set of potential subsets. This introduces the flexibility to select not only the best subset but also the second-best, third-best, and so on. To illustrate, to choose the second-best defined subset, one can initiate from the root state  and sequentially opt for the action corresponding to the second-largest estimated edge flow.

\begin{figure}[ht] 
\centering
 \vspace{-1mm}
{\label{fig:lossideal}\includegraphics[width = 9.5cm]{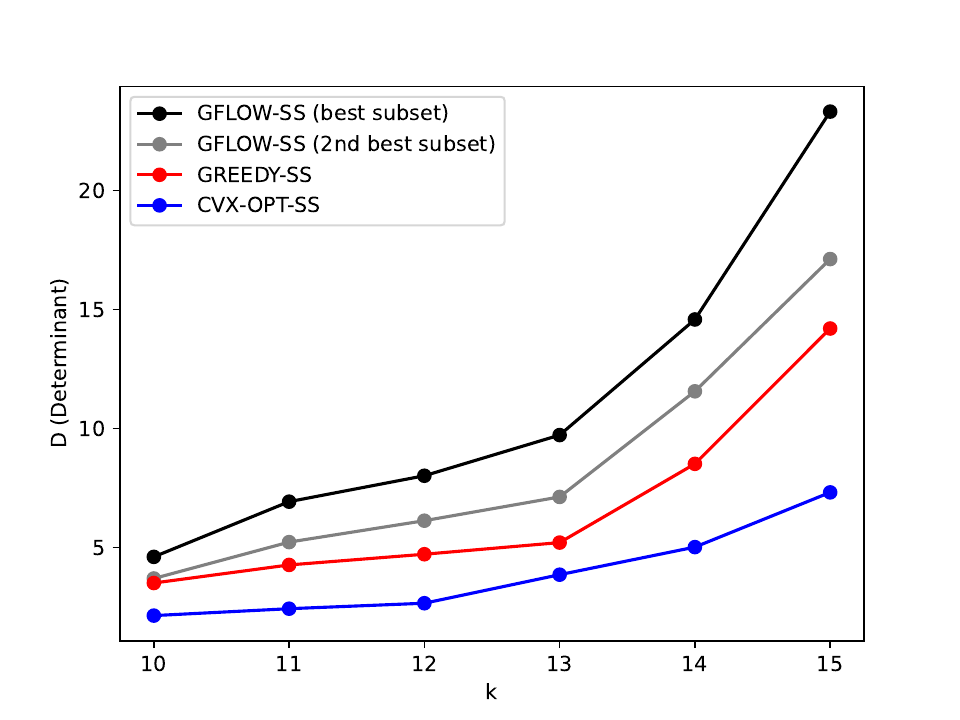}}%
\vspace{-0.5cm}
\caption{\small Comparison of the performance of \textbf{GFLOW-SS} (both for the best and the $2$nd best subset), \textbf{GREEDY-SS} and \textbf{CVX-OPT-SS} for the problem of selecting $k$ sensors out of $50$. The parameter $k$ ranges from $10$ to $15$. Each point corresponds to the average performance of the respective approach over $8$ different instantiations. In order to select the $2$nd best subset, the action that corresponds to the $2$nd best flow is chosen at each state. The \textbf{GFLOW-SS} approach is trained for $40000$ root-to-leaf MDP trajectories for every instantiation of the problem. This corresponds to $40000 \times k$ gradient descent steps.} \label{compare linear estimation 2}
\vspace{-2mm}
\end{figure}

Fig. \ref{compare linear estimation 2} presents a comparison of the performance between the best and the second-best subsets obtained from the \textbf{GFLOW-SS} approach, alongside the performances of \textbf{GREEDY-SS} and \textbf{CVX-OPT-SS}. The scenario involves selecting $k$ sensors out of a total of $m=50$, where $k$ ranges from $10$ to $15$. As illustrated in the figure, the second-best subset consistently outperforms both \textbf{GREEDY-SS} and \textbf{CVX-OPT-SS} approaches across all depicted values of $k$.

\section{Sensor Selection in ISAC}

The integration of sensing capabilities into communications stands as a pivotal direction in the evolution of the next generation of wireless systems \cite{bourdoux20206g}. This convergence leads to the contemporary framework of ISAC systems \cite{10188491}. Antenna selection emerges as a viable solution aimed at curbing power consumption and system costs in ISAC systems, all while upholding adequate performance for both sensing and communication modules. In \cite{9747651}, a supervised learning approach is proposed to select active antennas and optimize the transmit beamforming matrix, aiming to optimize a linear combination of the  CRB of target estimates and the overall communication rate. Additionally, \cite{10097184} examines an ISAC setting with quantized phase shifters and introduces a deep reinforcement learning approach to dynamically select a subset of antennas for transmission in each channel, thereby mitigating antenna coupling. 

\subsection{System Model}
We adopt the system model of \cite{9747651}. We consider a hybrid beamforming MIMO ISAC system that possesses $N_{t}$ transmit antennas, $N_{r}$ receive antenna elements and $N_{s}$ RF chains. The system performs two tasks: it is actively tracking a distributed target and is communicating with a single user which is equipped with $N_{s}$ antennas. The transmitter performs hybrid beamforming with beamforming matrix $\mathbf{F} \in \mathbb{C}^{N_{t} \times N{s}}$. The transmitted signal is:
\begin{equation}
    \mathbf{V} = \mathbf{F} \mathbf{X},
\end{equation}
where $\mathbf{X} \in \mathbb{C}^{N_{s} \times N_{t}}$ consists of $N_{s}$ unit power streams of length $L$, which translates to $\frac{1}{L} \mathbb{E}\left[ \mathbf{X} \mathbf{X}^{H}\right] = \mathbf{I}_{N_{s}}$. Furthermore, the channel matrix is denoted as $\mathbf{H} \in \mathbb{C}^{N_{s} \times N_{t}}$. The objective is to choose $N_{s}$ out of $N_{t}$ transmission antenna elements. To achieve this, a selection matrix $\mathbf{S} \in \{0,1\}^{N_{s} \times N_{t}}$ is introduced. Initially, all elements of $\mathbf{S}$ are set to $0$. Subsequently, each index and row are processed sequentially. During processing, each row of $\mathbf{S}$ remains $0$ except for the corresponding index of the selected element, which is set to $1$. Consequently, the matrix $\mathbf{S}^{H}\mathbf{S}$ is a diagonal matrix where each diagonal element is $0$ if the corresponding antenna is not selected, or $1$ if the corresponding antenna element is selected to be active. 

The CRB of the radar estimates of the sparse array is \cite{9747651}:
\begin{equation}
    \text{CRB}(\mathbf{S}, \mathbf{F}) = \frac{N_{r}}{L} \text{trace}\left((\mathbf{S} \mathbf{F} \mathbf{F}^{H} \mathbf{S}^{H})^{-1}\right).
\end{equation}
Correspondingly, the communication rate of the selected sensors is \cite{9747651}:
\begin{equation}
    G_{\text{rate}}(\mathbf{S}, \mathbf{F}) = \text{log}_{2}| \mathbf{I}_{N_{r}} + \frac{1}{N_{s}} \mathbf{H}\mathbf{S}^{H} \mathbf{S} \mathbf{F}\mathbf{F}^{H}\mathbf{S}^{H} \mathbf{S} \mathbf{H}^{H}|.
\end{equation}
The objective is to select the antenna elements (therefore implicitly select the selection matrix $\mathbf{S}$) and the beamforming matrix $\mathbf{F}$ such that a convex combination of the communication rate and the inverse of the CRB is maximized.

\begin{equation}
\begin{aligned}
& \text{maximize}_{\mathbf{S}, \mathbf{F}}
& & \beta_{\text{CRB}} \frac{1}{\text{CRB}(\mathbf{S}, \mathbf{F})} +  \beta_{\text{rate}}  G_{\text{rate}}(\mathbf{S}, \mathbf{F}) \\
& \text{subject to}
& & \mathbf{S} \text{ is selection matrix, selecting } N_{s} \text{ out of } N_{t}
\end{aligned}
\label{DFRC objective}
\end{equation}

We denote  as $\boldsymbol{\beta} = [ \beta_{\text{CRB}},  \beta_{\text{rate}} ]$ the $2$-dimensional vector of the coefficients of the two terms of the objective. Both scalar parameters should be nonnegative, $\beta_{\text{CRB}} \geq 0$, $\beta_{\text{rate}} \geq 0$. Furthermore they should add to 1, $\beta_{\text{CRB}} + \beta_{\text{rate}} = 1 $. We refer to $\boldsymbol{\beta}$ as the preference vector for the rest of the manuscript.

\subsection{A Multiobjective GFlowNet Approach}

The proposed GFlowNet method, outlined in Section \ref{general method}, readily extends to the specific ISAC scenario. Specifically, each terminal state $\mathbf{x}$ of a Sensor Selection MDP can be directly mapped to a selection matrix $\mathbf{S}$. If the selected antenna element positions (i.e., the indices of the 1 elements in $\mathbf{x}$) are $l_{1}, \dots, l_{N_{s}}$, then the corresponding selection matrix $\mathbf{S}(\mathbf{x})$ is defined as follows:

\begin{equation}
    s_{ij} = \begin{cases}
1 & \text{if } l_{i} = j \\
0 & \text{if } \text{otherwise}
\end{cases}
\end{equation}

Given a terminal state of the MDP and a fixed valid preference vector $\boldsymbol{\beta}$, the reward of the terminal state can be computed as:
\begin{equation}
    R(\mathbf{x}) = \beta_{\text{CRB}} \frac{1}{\text{CRB}(\mathbf{S}(\mathbf{x}), \mathbf{F})} +  \beta_{\text{rate}}  G_{\text{rate}}(\mathbf{S}(\mathbf{x}), \mathbf{F}).
\end{equation}
Considering that both the CRB and the communication rate depend not only on the selected antennas but also on the precoding matrix $\mathbf{F}$, upon reaching the terminal state $\mathbf{x}$, we can derive the corresponding optimal precoding matrix $\mathbf{F}^{*}$ by maximizing the objective function outlined in eq. \eqref{DFRC objective} via a fixed number of gradient descent steps. Given that the convex combination of the inverse CRB and the communication rate is a function of the complex matrix $\mathbf{F}$, the Wirtinger conjugate derivative \cite{wei2017conjugate} can be utilized. Hence, the reward function of the MDP is defined as follows:
\begin{equation}
    R(\mathbf{x}) = \beta_{\text{CRB}} \frac{1}{\text{CRB}(\mathbf{S}(\mathbf{x}), \mathbf{F}^{*})} +  \beta_{\text{rate}}  G_{\text{rate}}(\mathbf{S}(\mathbf{x}), \mathbf{F}^{*}).
\end{equation}

\subsection{Multiobjective Formulation}
The limitation of directly employing the proposed GFlowNet method in the context of the ISAC system is that it is applicable only for an appriori determined vector $\boldsymbol{\beta}$. Should there be any alterations to the coefficient terms, the reward function would consequently change, necessitating the problem to be resolved anew. In this subsection, we present an alternative formulation that tackles the challenge of multiobjective optimization, thereby circumventing the necessity of retraining the GFlowNet for different values of the preference vector, each corresponding to distinct design system specifications.

The flow-matching objective outlined in eq. \eqref{flow matching} pertains to the direct parameterization of the optimal flow within the relevant MDP. Along the lines of \cite{malkin2022trajectory} , we introduce three additional quantities derived from the optimal flow $F$: the forward sampling policy denoted as $P^{F}(\mathbf{s}^{'} = T(\mathbf{s}, \mathbf{a}) | \mathbf{s}) = \frac{F(s, a)}{F(s)}$, the backward sampling policy represented by $P^{B}(\mathbf{s} | \mathbf{s}^{'} = T(\mathbf{s}, \mathbf{a}) ) = \frac{F(\mathbf{s}, \mathbf{a})}{F(\mathbf{s}^{'})}$, and the partition function of the MDP, denoted as $Z = \sum_{\mathbf{x} \in \mathcal{X}} R(\mathbf{x})$.

We then use a function approximator, such as a neural network, for the forward policy $P_{\mathbf{w}}^{F}(\mathbf{s}^{'} | \mathbf{s})$, a function approximator for the backward policy $P_{\mathbf{w}}^{B}(\mathbf{s} | \mathbf{s}^{'})$, and a learnable parameter for the partition function $Z_{\mathbf{w}}$. Then, for a root-to-leaf trajectory $\tau = (\mathbf{s}_{0} \rightarrow \mathbf{s}_{1} \rightarrow \dots \rightarrow \mathbf{s}_{n} = \mathbf{x})$, the trajectory balance objective is:
\begin{equation}
    L_{\mathbf{w}}^{TB}(\tau) = \left( \text{log} \frac{Z_{\mathbf{w}} \prod_{t=1}^{n} P_{\mathbf{w}}^{F} (\mathbf{s}_{t} | \mathbf{s}_{t-1}) }{R(\mathbf{x}) \prod_{t=1}^{n} P_{\mathbf{w}}^{B} (\mathbf{s}_{t-1} | \mathbf{s}_{t})}\right)^{2}.
    \label{trajectory balance loss}
\end{equation}

By identifying a parameter vector $\mathbf{w}^{*}$ that minimizes the aforementioned objective across all root-to-leaf trajectories, sampling actions according to the converged forward policy $P_{\mathbf{w}^{*}}^{F}$ guarantees that terminal states are sampled proportionally to their corresponding rewards.

The trajectory balance variation of the GFlowNet learning objective directly parameterizes the partition function of the MDP, thereby enabling the parametrization of a range of action-sampling distributions corresponding to various reward functions within the MDP's structure. These reward functions can be conditioned on multiple variables \cite{jain2023multi}. In the context of sensor selection for ISAC, the conditioning variable for the reward is the preference vector $\boldsymbol{\beta}$ representing the coefficients of the two objective terms. Hence, the reward of a terminal state $\mathbf{x}$ also depends on the choice of $\boldsymbol{\beta}$, denoted as $R(\mathbf{x}, \boldsymbol{\beta})$. Consequently, the estimations of forward and backward policies are influenced by $\boldsymbol{\beta}$ as well, expressed as $P_{\mathbf{w}}^{B} (\mathbf{s}_{t-1} | \mathbf{s}_{t} ; \boldsymbol{\beta})$ and $P_{\mathbf{w}}^{F} (\mathbf{s}_{t} | \mathbf{s}_{t-1}; \boldsymbol{\beta})$. In this more generalized formulation, the partition function is not a single learnable scalar but rather a function approximator that takes the preference vector $\boldsymbol{\beta}$ as input, $Z_{\mathbf{w}}(\boldsymbol{\beta})$. 

The main difference between the GFlowNet approach that is outlined in Section \ref{general method} and the multiobjective GFlowNet approach, besides the form of the objective, is that before sampling each root-to-leaf trajectory for gradient descent, a $\boldsymbol{\beta}$ vector is sampled from a set of predefined values.
{ 
\begin{algorithm}
\caption{\textbf{MOGFLOW - SS}}
\label{mogflow algo}
\begin{algorithmic}
\State{Initialize $P_{\mathbf{w}}^{B} (\mathbf{s}^{'} | \mathbf{s} ; \boldsymbol{\beta})$,$P_{\mathbf{w}}^{F} (\mathbf{s} | \mathbf{s}^{'}; \boldsymbol{\beta})$, $Z_{\mathbf{w}}(\boldsymbol{\beta})$,  $\zeta \in [0,1]$ for exploration,  learning rate $\eta$, dictionary of $\boldsymbol{\beta}$ values $\{\boldsymbol{\beta}^{1}, \dots, \boldsymbol{\beta}^{d} \}$, $N_{\text{Wir}}$ the number of Wirtinger conjugate gradient steps for optimizing the beamforming matrix $\mathbf{F}$}
\For {\texttt{all root-to-leaf trajectories}}
    \State{$\mathbf{s} = \mathbf{s}_{0} = [0]^{N_{t}}$}
    \State{Sample $\boldsymbol{\beta}$ randomly from the dictionary}
    \State{Compute $Z_{\mathbf{w}}(\boldsymbol{\beta})$}
    \For{$N_{s}-1$ transitions}
        \State {Sample $z \sim \mathcal{U}(0,1)$ (Uniform distribution)}
        \State {If $z < \zeta$ choose $\mathbf{s}^{'}$ randomly}
        \State {If $z >= \zeta$ sample $\mathbf{s}^{'}$ from $P_{\mathbf{w}}^{F} (\mathbf{s} | \mathbf{s}^{'}; \boldsymbol{\beta})$}
        \State {Compute $P_{\mathbf{w}}^{B} (\mathbf{s}^{'} | \mathbf{s} ; \boldsymbol{\beta})$}
        \State { $\mathbf{s}$ = $\mathbf{s}'$}
    \EndFor
        \State {Sample $z \sim \mathcal{U}(0,1)$}
        \State {If $z < \zeta$ choose choose $\mathbf{s}^{'}$ randomly randomly}
        \State {If $z >= \zeta$ choose $\mathbf{x}$ from $P_{\mathbf{w}}^{F} (\mathbf{s} | \mathbf{x}; \boldsymbol{\beta})$}
        \State {Compute $P_{\mathbf{w}}^{B} (\mathbf{x} | \mathbf{s} ; \boldsymbol{\beta})$}
        \State{Compute $\mathbf{F}^{*}$ by $N_{\text{Wir}}$ gradient steps and learning rate $\eta$}
        \State{Compute $R(\mathbf{x}, \boldsymbol{\beta})$}
        \State{$\tau = (\mathbf{s}_{0} \rightarrow \mathbf{s}_{1} \rightarrow \dots \rightarrow \mathbf{s}_{N_{s}} = \mathbf{x}) $}
        \State {$\mathbf{w}' \rightarrow \mathbf{w} - \eta \nabla_{\mathbf{w}} L_{\mathbf{w}}^{TB}(\tau)$}
\EndFor
\end{algorithmic}
\end{algorithm}
}

The multiobjective reformulation of the GFlowNet approach for sensor selection in the ISAC system is referred to as \textbf{MOGFLOW-SS}, and its procedural framework is outlined in Algorithm \ref{mogflow algo}. A key practical advantage of \textbf{MOGFLOW-SS} over \textbf{GFLOW-SS} lies in its ability to learn a spectrum of sampling distributions conditioned not only on the current state of the MDP but also on the design choice of the preference vector $\boldsymbol{\beta}$. This capability facilitates generalization across values of the preference vector that may not have been encountered during the training phase.

\subsection{Experiments}

We consider a practical setting where the number of transmit antennas is $N_{t} = 80$ and the number of receive antennas is also $N_{r} = 80$. The number of RF chains is $N_{s} = 10$. The length of the ISAC symbol is chosen to be $L = 100$. Every element of the channel matrix $\mathbf{H}$ is sampled from a standard complex normal distribution $H_{ij} \sim \mathcal{C}\mathcal{N}(0,1)$. 

The backward policy $P^{B} (\mathbf{s}^{'} | \mathbf{s} ; \boldsymbol{\beta})$ serves as a prior over the parent nodes for each MDP state. It is essentially an artifact of the training process and is discarded post-convergence. While it could be explicitly parameterized, following the methodology outlined in \cite{malkin2022trajectory}, we opt for a uniform distribution over the potential parent states of a given state. Specifically, for every state in the Sensor Selection MDP (excluding the root), the backward policy is selected to be inversely proportional to the number of ``1"s present in the vector representation of the state,  $P^{B} (- | \mathbf{s} ; \boldsymbol{\beta}) = 1 / || \mathbf{s}||_{0}$.

\begin{figure}[ht] 
\centering
 \vspace{-1mm}
{\label{fig:lossideal}\includegraphics[width = 9.0cm]{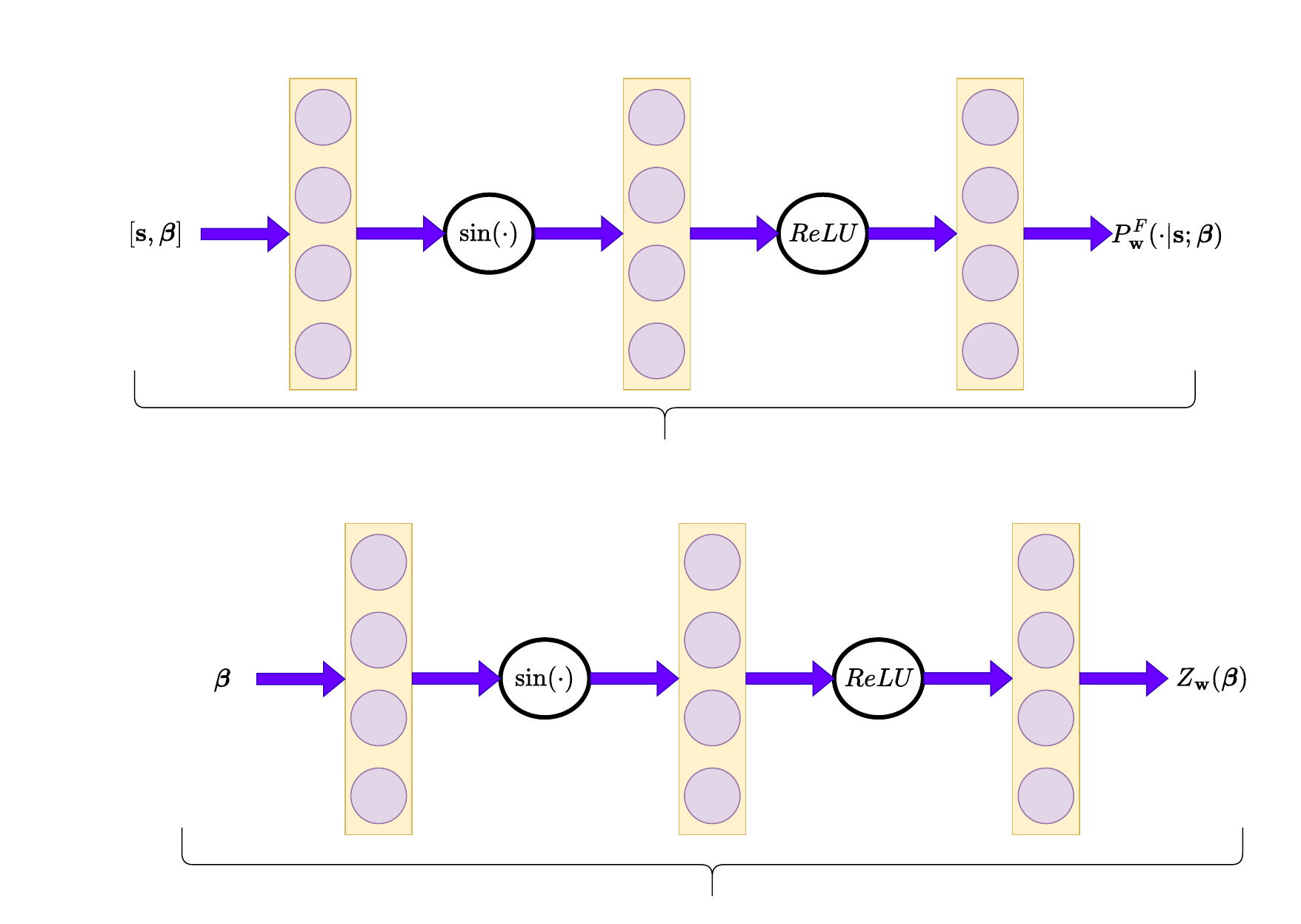}}%
\vspace{-0.5cm}
\caption{\small The architecture of the two parametrizations ($Z_{\mathbf{w}}(\boldsymbol{\beta})$, $P_{\mathbf{w}}^{F} (\cdot| \mathbf{s}; \boldsymbol{\beta})$) employed for \textbf{MOGFLOW-SS}.} \label{parametrizations}
\end{figure}

The forward policy is realized through a $3$-layer dense neural network, each layer comprising $150$ neurons. Remarkably, the first layer incorporates the Fourier kernel, as introduced by \cite{9676432}, initialized with elements drawn from a zero-mean Gaussian distribution with a variance of $0.1$. Similarly, the partition function $Z_{\mathbf{w}}(\boldsymbol{\beta})$ is parameterized in the log-domain, employing a $3$-layer neural network structure identical to that of the forward policy. The variance of the initialization for the Fourier kernel elements is set to $0.001$. Despite employing two distinct parametrizations, both are referenced using the same parameter vector $\mathbf{w}$, facilitating joint optimization of both networks with a consistent objective and optimizer. Specifically, we utilize the Adam optimizer \cite{kingma2014adam} with a learning rate of $10^{-5}$.

Regarding the preference vector $\boldsymbol{\beta} = [ \beta_{\text{CRB}}, \beta_{\text{rate}}]$, where both coefficients are nonnegative and sum up to $1$, we opt to represent them using a single scalar variable $n$. Specifically, we set $\beta_{\text{CRB}} = n$ and $\beta_{\text{rate}} = 1 - n$, where $n \in (0,1)$. For the training process, we adopt a strategy of updating on three distinct values of $\boldsymbol{\beta}$. Each value corresponds to a different value of $n$, namely ${0.1, 0.5, 0.9}$.

In formulating the reward function for a given terminal state-subset $\mathbf{x}$ and a preference variable $n$, we encounter the challenge that the two objective terms span different value ranges. To address this discrepancy, we introduce an additional scalar parameter $c_{\text{scale}}$, which we use to scale the inverse of the CRB term. This scaling ensures that both objective terms exhibit roughly comparable ranges of values. To determine the parameter $c_{\text{scale}}$, we compute the product $\text{CRB}(\mathbf{S}(\mathbf{x}), \mathbf{F}^{}) \times G_{\text{rate}}(\mathbf{S}(\mathbf{x}), \mathbf{F}^{})$ for $10$ randomly sampled subsets of size $N_{s}$. Finally, the reward function is formulated as follows:
\begin{equation}
    R(\mathbf{x}, n) = n \frac{c_{\text{scale}}}{\text{CRB}(\mathbf{S}(\mathbf{x}), \mathbf{F}^{*})} +  (1-n)  G_{\text{rate}}(\mathbf{S}(\mathbf{x}), \mathbf{F}^{*}),
\end{equation}
where $c_{\text{scale}}$ is chosen to be $20000$ for the setting of interest.

The training process encompasses a total of $60,000$ episodes (root-to-leaf trajectories) and involves $N_{\text{Wir}} = 50$ Wirtinger conjugate gradient steps for estimating $\mathbf{F}^{*}$ at the conclusion of each episode.
\begin{figure}[ht] 
\centering
 \vspace{-1mm}
{\label{fig:lossideal}\includegraphics[width = 9.0cm]{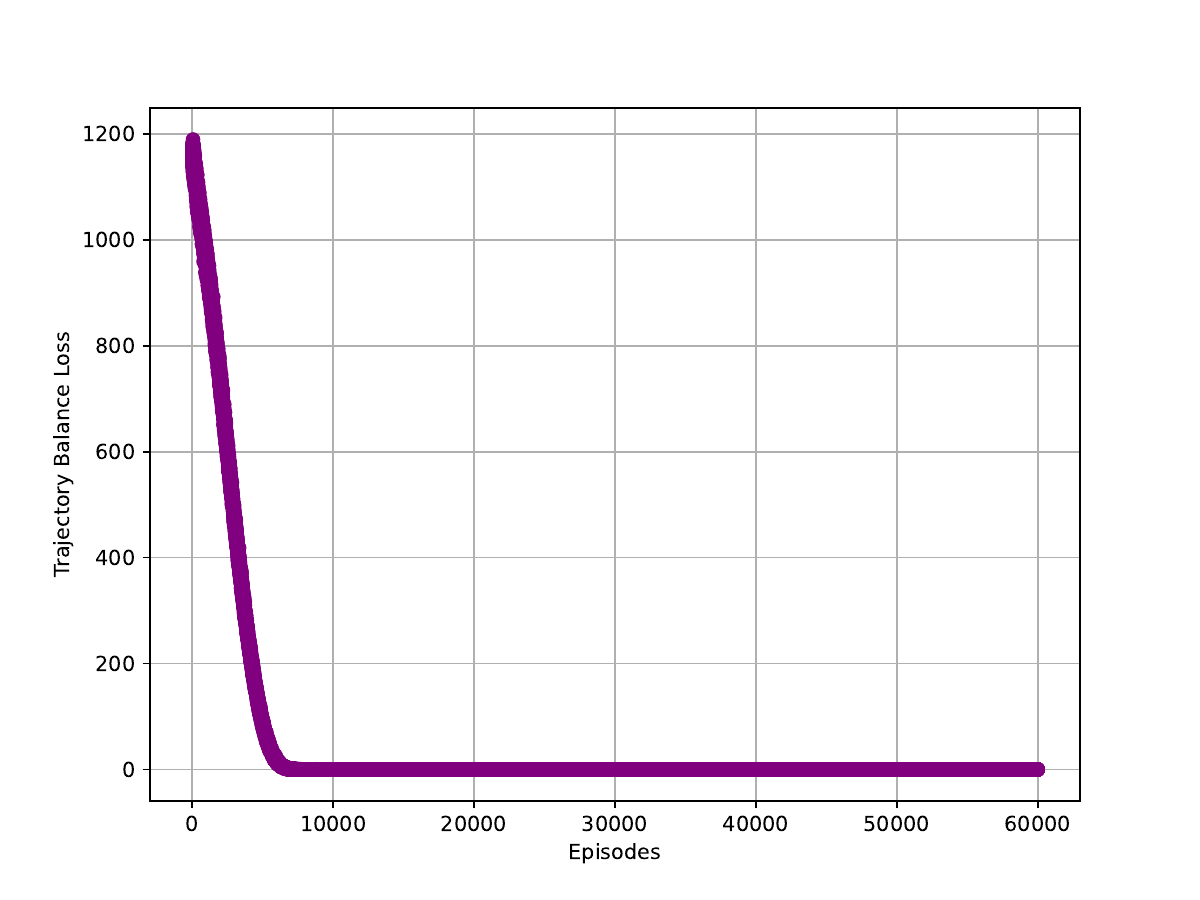}}%
\vspace{-0.5cm}
\caption{\small The trajectory balance loss for \textbf{MOGFLOW-SS} is computed over $60000$ episodes, representing root-to-leaf trajectories.} \label{tb_loss}
\vspace{-2mm}
\end{figure}

Fig. \ref{tb_loss} depicts the trajectory balance loss (eq. \ref{trajectory balance loss}) over the course of training. The plot illustrates a rapid decrease in the loss value, ultimately converging to values very close to zero. This convergence signifies that the forward policy effectively captures the action-sampling policy, ensuring that terminal states are sampled proportionally to the performances of their respective subsets.

\begin{figure}[ht] 
\centering
 \vspace{-1mm}
{\label{fig:lossideal}\includegraphics[width = 9.0cm]{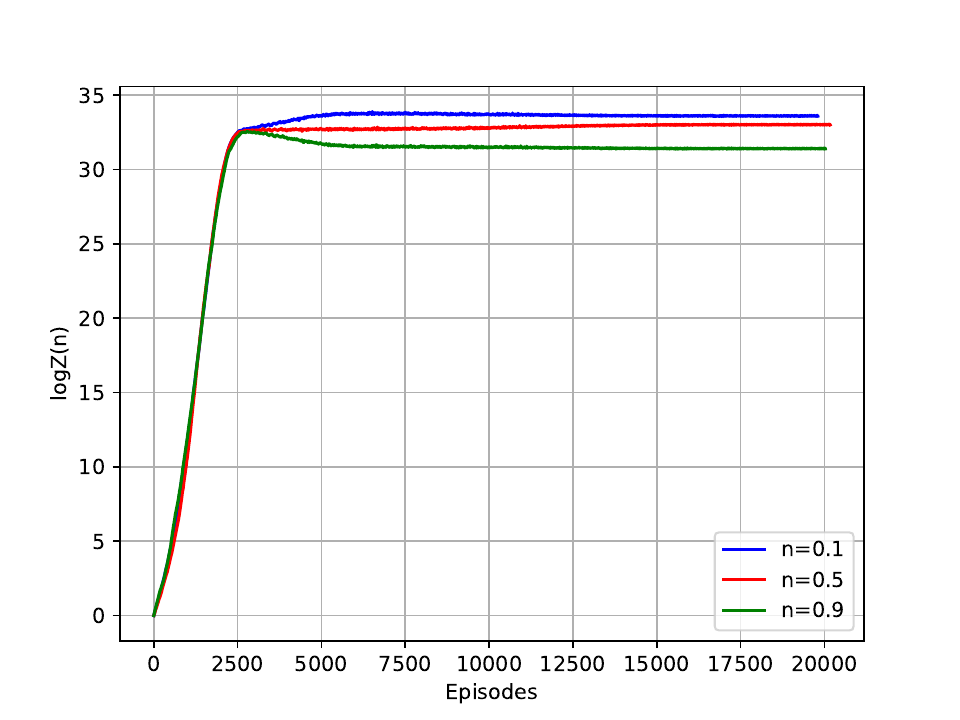}}%
\vspace{-0.5cm}
\caption{\small The value of  $\text{log}Z$ during training for the $3$ different values of $n$.} \label{logz}
\vspace{-2mm}
\end{figure}

Fig. \ref{logz} illustrates the evolution of the partition function (in the log domain) throughout the training process, considering all three values of $n$ ($0.1$, $0.5$, and $0.9$). Each $n$ value corresponds to a distinct preference vector $\boldsymbol{\beta}$. While the partition function typically remains constant for a single MDP \cite{bengio2023gflownet}, in this scenario, it varies with $n$. Consequently, as each $n$ signifies a different reward function, the final values of $\text{log}Z$ upon convergence differ across different $n$ values. Lower values of $n$ prioritize the objective term linked to the communication system (communication rate), whereas higher values of $n$ emphasize the term associated with radar performance (the inverse of the CRB). This nuance allows for flexible optimization tailored to diverse system requirements.

Upon completing the training phase, we select five distinct values of $n$: $0.1$, $0.3$, $0.5$, $0.7$, and $0.9$. Notably, three of these ($0.1$, $0.5$, $0.9$) were employed during training, while the remaining two ($0.3$, $0.7$) were not utilized for training purposes. For each $n$ value, we implement a greedy strategy to sample the terminal state, selecting the action that maximizes the estimated forward policy network at each state, starting from the root. This approach results in a subset containing $N_{s}$ active antenna elements for each $n$ value.

\begin{figure}[ht] 
\centering
 \vspace{-1mm}
{\label{fig:lossideal}\includegraphics[width = 9.0cm]{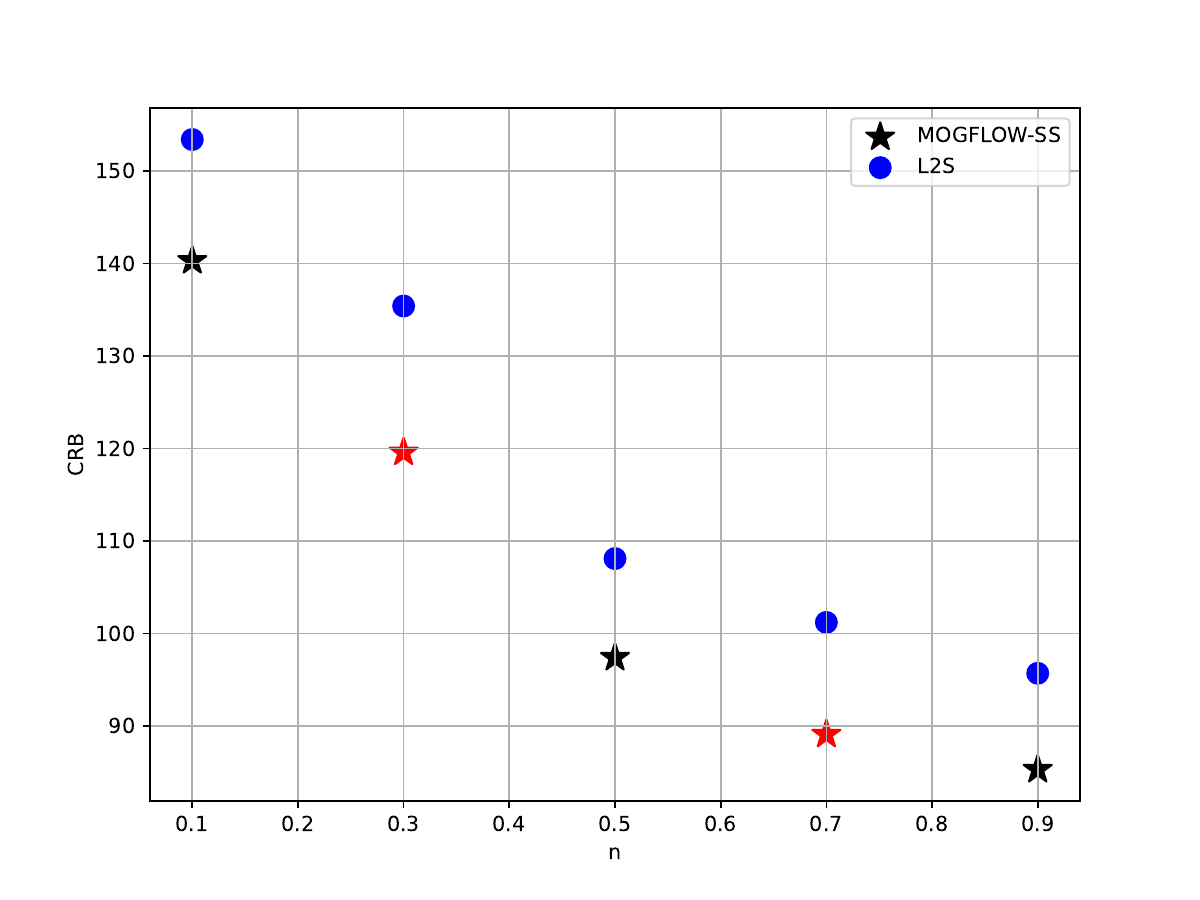}}%
\vspace{-0.5cm}
\caption{\small  The CRB values (lower values indicate better performance) associated with subsets selected by both the \textbf{MOGFLOW-SS} and the \textbf{L2S} methods for five different values of $n$. Stars represent the performance of subsets recovered by \textbf{MOGFLOW-SS}, while circles represent subsets recovered by \textbf{L2S}. \textbf{Black} stars denote values of $n$ used during training, whereas \textbf{\textcolor{red}{Red}} stars denote values of $n$ not included in the training process. Every point in the plot corresponds to the average over $15$ different seeds.} \label{crb fig}
\vspace{-2mm}
\end{figure}

Fig. \ref{crb fig} illustrates the CRB values associated with the optimal sensor subset and beamforming matrix $\mathbf{F}^{*}$ obtained through the \textbf{MOGFLOW-SS} method, across the five selected values of $n$. This plot provides a comparison between the CRB performance achieved by \textbf{MOGFLOW-SS} and the method presented in the work of \cite{xu2022cramer}, denoted as \textbf{L2S} in their original paper. Notably, the approach of \cite{xu2022cramer} examines the same setting. The results demonstrate \textbf{MOGFLOW-SS}'s clear superiority over \textbf{L2S}. Consistently, \textbf{MOGFLOW-SS} yields subsets and corresponding beamforming matrices that lead to lower CRB values across all $n$ values. Here, lower CRB values obtained through \textbf{MOGFLOW-SS} signify enhanced radar localization performance.

\begin{figure}[ht] 
\centering
 \vspace{-1mm}
{\label{fig:lossideal}\includegraphics[width = 9.0cm]{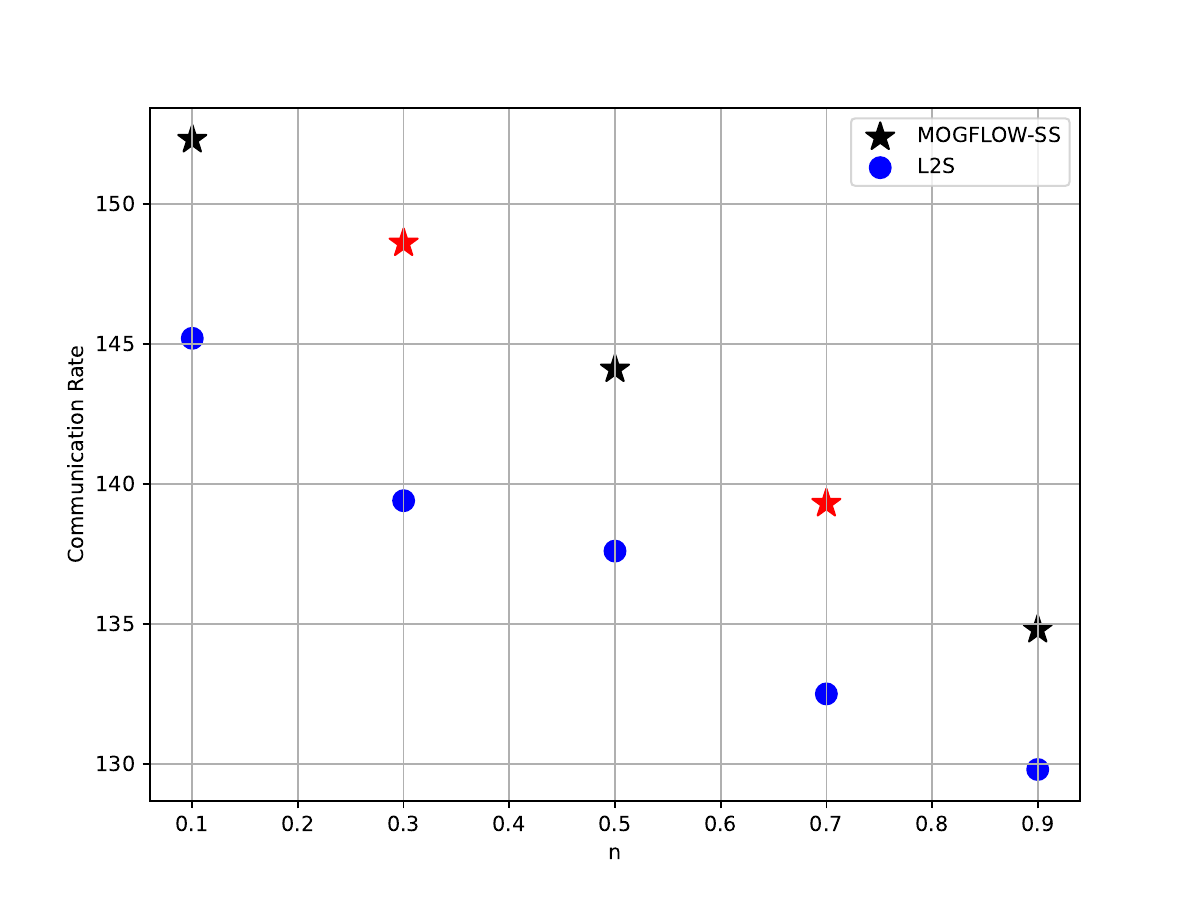}}%
\vspace{-0.5cm}
\caption{\small  The communication rate values (higher values indicate better performance) associated with subsets selected by both the \textbf{MOGFLOW-SS} and the \textbf{L2S} methods for five different values of $n$. Stars represent the performance of subsets recovered by \textbf{MOGFLOW-SS}, while circles represent subsets recovered by \textbf{L2S}. \textbf{Black} stars denote values of $n$ used during training, whereas \textbf{\textcolor{red}{Red}} stars denote values of $n$ not included in the training process. Every point in the plot corresponds to the average over $15$ different seeds.} \label{com fig}
\vspace{-2mm}
\end{figure}

Fig. \ref{com fig} illustrates the communication rate values corresponding to the optimal sensor subset and beamforming matrix $\mathbf{F}^{*}$ obtained through \textbf{MOGFLOW-SS} across five selected values of $n$. The communication rate performance achieved by \textbf{MOGFLOW-SS} is compared with that of subsets derived from \textbf{L2S}. Notably, \textbf{MOGFLOW-SS} surpasses \textbf{L2S} by generating subsets and corresponding beamforming matrices associated with higher communication rates across all $n$ values.

The parameter $n$ serves as a coefficient that scales the inverse of the CRB term in the reward function. With the coefficients of both reward terms summing to $1$, higher $n$ values prioritize the CRB term, while lower values prioritize the communication rate term.
This distinction is crucial in understanding the trends observed in the derived subsets by \textbf{MOGFLOW-SS}. As $n$ increases, emphasizing the CRB term, \textbf{MOGFLOW-SS} produces subsets with smaller CRB values and correspondingly larger inverse CRB values, as depicted in Fig. \ref{crb fig}. Conversely, larger $n$ values result in subsets with lower communication rates, as observed in Fig. \ref{com fig}.
Moreover, for each $n$ value, \textbf{MOGFLOW-SS} consistently outperforms \textbf{L2S} in terms of both radar performance (lower CRB values) and communication performance (higher communication rates). This superiority is particularly notable considering that \textbf{L2S} requires training from scratch for each $n$ value, whereas \textbf{MOGFLOW-SS} is trained once for all $n$ values. Despite this, \textbf{MOGFLOW-SS} effectively balances the trade-off between radar and communication performance, making it superior in solving the selection problem across varying $n$ values. 

The pivotal aspect of \textbf{MOGFLOW-SS} performance is elucidated by the red points in Figures \ref{crb fig} and \ref{com fig}. Specifically, these red points represent the performance of \textbf{MOGFLOW-SS} for the two $n$ values excluded from the training process. Essentially serving as a test set, these values validate the model's generalization capability beyond its training data.
Remarkably, the trained GFlowNet exhibits robust generalization to preference variable values absent during gradient descent updates. This underscores the method's ability to effectively adapt to the entire spectrum of system configurations without necessitating additional training or resources, thereby bolstering its practical applicability.

\subsubsection{Thorough Assessment of Generalization}

While the performance of \textbf{MOGFLOW-SS} on unseen values of $n$ (specifically $n=0.3$ and $n=0.7$) highlights its generalization capability, questions arise regarding its adaptability to diverse system settings and requirements. Notably, each subset's performance depends not only on the active sensor selection but also on the estimation of the beamforming matrix $\mathbf{F}^{}$. Since $\mathbf{F}^{}$ is optimized via gradient descent on the objective function, different $n$ values may yield varying CRB and communication rate outcomes for the same subset. Consequently, distinct $n$ values may result in different $\mathbf{F}^{*}$ values, even for identical terminal states $\mathbf{x}$. Therefore, genuine generalization would entail \textbf{MOGFLOW-SS} selecting different subsets for different $n$ values, particularly for those ($0.3$ and $0.7$) absent during the training phase.

\begin{table}[htbp]
    \centering
    \caption{Antennas chosen by \textbf{MOGFLOW-SS} for \underline{different} values of $n$}
    \label{tab:positions}
    \begin{tabular}{c | >{\centering\arraybackslash}m{6cm}}
        \hline
        $n$ & Indices of Antenna Positions Chosen \\ \hline
        $0.1$ & [$6$, $12$, $14$, $15$, $18$, $32$, $45$, $46$, $62$, $77$] \\ \hline
        $\mathbf{0.3}$ & [$\mathbf{6}$, $\mathbf{14}$, $\mathbf{35}$, $\mathbf{38}$, $\mathbf{46}$, $\mathbf{53}$, $\mathbf{57}$, $\mathbf{62}$, $\mathbf{66}$, $\mathbf{68}$] \\ \hline
        $0.5$ & [$0$, $4$, $10$, $27$, $33$, $38$, $39$, $51$, $60$, $68$] \\ \hline
        $\mathbf{0.7}$ & [$\mathbf{0}$, $\mathbf{10}$, $\mathbf{19}$, $\mathbf{22}$, $\mathbf{29}$, $\mathbf{33}$, $\mathbf{38}$, $\mathbf{40}$, $\mathbf{51}$, $\mathbf{68}$] \\ \hline
        $0.9$ & [$0$, $10$, $19$, $29$, $33$, $38$, $40$, $51$, $60$, $67$] \\ \hline
    \end{tabular}
    \label{table indices}
\end{table}

Table \ref{table indices} displays the indices of the selected antennas, determined by \textbf{MOGFLOW-SS}, across different $n$ values for a single random seed, representing a sole iteration of the training process. Notably, while significant overlap exists among the antenna subsets, each subset corresponding to a specific $n$ value remains distinct. This distinctiveness is particularly pronounced for $n=0.3$ and $n=0.7$, which were excluded from the training process. The uniqueness of each subset underscores the robust generalization capacity inherent in the GFlowNet paradigm, facilitating the accurate estimation of well-performing antenna configurations across the entire range of $n \in (0,1)$. Each $n$ value embodies a distinct trade-off between the radar and communication systems. Notably, this uniqueness persists across all $15$ seeds utilized in the experimentation.

\section{Remarks}
The challenge of sensor selection problems lies in the combinatorial explosion of the solution set. While greedy selection methods offer efficiency by narrowing down the search space, they often yield optimal or near-optimal solutions only under highly specific problem instances. Motivated by this limitation, our work seeks to pioneer a paradigm that can reason about the quality of a vast majority of potential solutions without the need for exhaustive evaluation.
Initially, the idea of modeling selection as an MDP may seem counterintuitive, as it ostensibly expands the search space. This arises from the fact that the number of possible state-action pairs in the sensor selection MDP exceeds the number of possible subsets (terminal states). However, this approach enables us to leverage GFlowNets to implicitly parameterize the action sampling distribution. By doing so, we transition the search space from discrete subsets or state-action pairs to the continuous parameter space of the function approximator class.
Operating within this framework, we can employ stochastic gradient descent methods. Notably, the same parameter vector governs estimations for all state-action pairs, facilitating implicit generalization. Consequently, updates based on specific subsets influence estimations for all other subsets. This inherent generalization capability allows us to reason about subsets without the need for exhaustive evaluation.

In the multiobjective formulation, this generalization extends to the space of preference vectors. This aspect is enabled by the explicit parametrization of the partition function of the MDP (eq. \eqref{trajectory balance loss}).

\section{Conclusion}

The current study has investigated the broad domain of problems that are prevalent in signal processing and focus on the selection of a fixed-size subset of sensors from a deployed set in order to optimize a metric related to detection and estimation. These problems resemble knapsack-type scenarios, inheriting the combinatorial complexity stemming from the exponential growth of the solution set. Prior research has proposed numerous methodologies to address such problems, broadly categorized into two main groups: analytical approaches (such as convex optimization and greedy selection) and supervised machine learning methods.
Analytical approaches, while offering insights, are often reliant on specific properties of the objective function and tend to provide suboptimal solutions. On the other hand, supervised machine learning methods necessitate substantial amounts of annotated data for effective training. In light of these limitations, the current work has presented a novel general framework predicated on generative AI to tackle sensor selection problems, aiming to overcome the drawbacks of previous paradigms.
The proposed approach formulates the sensor selection problem as a deterministic MDP, where subsets of desired size emerge as terminal states. The objective is to learn an action sampling distribution that ensures that the cumulative probability of sampling a terminal state is proportional to the performance of the corresponding subset. Leveraging the deep generative modeling paradigm of GFlowNet, the proposed approach amortizes the cost of learning the action sampling distribution.
Empirical evaluation has demonstrated that the proposed approach outperforms both convex optimization methods and greedy selection approaches. Notably, the approach is trained on a significantly reduced subset of the solution set. Furthermore, a multiobjective formulation of the approach has been introduced and applied to a sensor selection problem for ISAC systems. This problem entails a trade-off between radar and communication system performance, encapsulated by a preference vector. The proposed approach adeptly manages this trade-off and demonstrates robust generalization even to preference vector values not used during the GFlowNet training process.

\bibliographystyle{IEEEtran}
\bibliography{main.bib}

\begin{thebibliography}{10}
\providecommand{\url}[1]{#1}
\csname url@samestyle\endcsname
\providecommand{\newblock}{\relax}
\providecommand{\bibinfo}[2]{#2}
\providecommand{\BIBentrySTDinterwordspacing}{\spaceskip=0pt\relax}
\providecommand{\BIBentryALTinterwordstretchfactor}{4}
\providecommand{\BIBentryALTinterwordspacing}{\spaceskip=\fontdimen2\font plus
\BIBentryALTinterwordstretchfactor\fontdimen3\font minus \fontdimen4\font\relax}
\providecommand{\BIBforeignlanguage}[2]{{%
\expandafter\ifx\csname l@#1\endcsname\relax
\typeout{** WARNING: IEEEtran.bst: No hyphenation pattern has been}%
\typeout{** loaded for the language `#1'. Using the pattern for}%
\typeout{** the default language instead.}%
\else
\language=\csname l@#1\endcsname
\fi
#2}}
\providecommand{\BIBdecl}{\relax}
\BIBdecl

\bibitem{li2008mimo}
J.~Li and P.~Stoica, \emph{MIMO radar signal processing}.\hskip 1em plus 0.5em minus 0.4em\relax John Wiley \& Sons, 2008.

\bibitem{4176505}
N.~H. Lehmann, A.~M. Haimovich, R.~S. Blum, and L.~Cimini, ``High resolution capabilities of mimo radar,'' in \emph{2006 Fortieth Asilomar Conference on Signals, Systems and Computers}, 2006, pp. 25--30.

\bibitem{4663892}
S.~Joshi and S.~Boyd, ``Sensor selection via convex optimization,'' \emph{IEEE Transactions on Signal Processing}, vol.~57, no.~2, pp. 451--462, 2009.

\bibitem{gondzio2012interior}
J.~Gondzio, ``Interior point methods 25 years later,'' \emph{European Journal of Operational Research}, vol. 218, no.~3, pp. 587--601, 2012.

\bibitem{6760894}
Y.~Wang, M.~Sznaier, and F.~Dabbene, ``A convex optimization approach to worst-case optimal sensor selection,'' in \emph{52nd IEEE Conference on Decision and Control}, 2013, pp. 6353--6358.

\bibitem{wielgus2024general}
A.~Wielgus and B.~Szlachetko, ``A general scheme of a branch-and-bound approach for the sensor selection problem in near-field broadband beamforming,'' \emph{Sensors}, vol.~24, no.~2, p. 470, 2024.

\bibitem{boyd2007branch}
S.~Boyd and J.~Mattingley, ``Branch and bound methods,'' \emph{Notes for EE364b, Stanford University}, vol. 2006, p.~07, 2007.

\bibitem{nosrati2017receiver}
H.~Nosrati, E.~Aboutanios, and D.~B. Smith, ``Receiver-transmitter pair selection in {MIMO} phased array radar,'' in \emph{2017 IEEE International Conference on Acoustics, Speech and Signal Processing (ICASSP)}.\hskip 1em plus 0.5em minus 0.4em\relax IEEE, 2017, pp. 3206--3210.

\bibitem{richard1991design}
M.~Richard and D.~Brian, ``Design of maximally sparse beamforming arrays,'' \emph{IEEE Trans. Antennas Propag}, vol.~39, no.~8, pp. 1178--1187, 1991.

\bibitem{6031934}
H.~Godrich, A.~P. Petropulu, and H.~V. Poor, ``Sensor selection in distributed multiple-radar architectures for localization: A knapsack problem formulation,'' \emph{IEEE Transactions on Signal Processing}, vol.~60, no.~1, pp. 247--260, 2012.

\bibitem{5717225}
M.~Shamaiah, S.~Banerjee, and H.~Vikalo, ``Greedy sensor selection: Leveraging submodularity,'' in \emph{49th IEEE Conference on Decision and Control (CDC)}, 2010, pp. 2572--2577.

\bibitem{majumder2023greedy}
K.~Majumder, S.~B. Pillai, and S.~Mulleti, ``Greedy selection for heterogeneous sensors,'' \emph{arXiv preprint arXiv:2307.00840}, 2023.

\bibitem{joung2016machine}
J.~Joung, ``Machine learning-based antenna selection in wireless communications,'' \emph{IEEE Communications Letters}, vol.~20, no.~11, pp. 2241--2244, 2016.

\bibitem{hearst1998support}
M.~A. Hearst, S.~T. Dumais, E.~Osuna, J.~Platt, and B.~Scholkopf, ``Support vector machines,'' \emph{IEEE Intelligent Systems and their applications}, vol.~13, no.~4, pp. 18--28, 1998.

\bibitem{elbir2019cognitive}
A.~M. Elbir, K.~V. Mishra, and Y.~C. Eldar, ``Cognitive radar antenna selection via deep learning,'' \emph{IET Radar, Sonar \& Navigation}, vol.~13, no.~6, pp. 871--880, 2019.

\bibitem{lecun1995convolutional}
Y.~LeCun, Y.~Bengio \emph{et~al.}, ``Convolutional networks for images, speech, and time series,'' \emph{The handbook of brain theory and neural networks}, vol. 3361, no.~10, p. 1995, 1995.

\bibitem{vu2021machine}
T.~X. Vu, S.~Chatzinotas, V.-D. Nguyen, D.~T. Hoang, D.~N. Nguyen, M.~Di~Renzo, and B.~Ottersten, ``Machine learning-enabled joint antenna selection and precoding design: From offline complexity to online performance,'' \emph{IEEE Transactions on Wireless Communications}, vol.~20, no.~6, pp. 3710--3722, 2021.

\bibitem{lin2021deep}
B.~Lin, F.~Gao, S.~Zhang, T.~Zhou, and A.~Alkhateeb, ``Deep learning-based antenna selection and {CSI} extrapolation in massive {MIMO} systems,'' \emph{IEEE Transactions on Wireless Communications}, vol.~20, no.~11, pp. 7669--7681, 2021.

\bibitem{diamantaras2021sparse}
K.~Diamantaras, Z.~Xu, and A.~Petropulu, ``Sparse antenna array design for {MIMO} radar using softmax selection,'' \emph{arXiv preprint arXiv:2102.05092}, 2021.

\bibitem{vaswani2017attention}
A.~Vaswani, N.~Shazeer, N.~Parmar, J.~Uszkoreit, L.~Jones, A.~N. Gomez, {\L}.~Kaiser, and I.~Polosukhin, ``Attention is all you need,'' \emph{Advances in neural information processing systems}, vol.~30, 2017.

\bibitem{xu2022cramer}
Z.~Xu, F.~Liu, and A.~Petropulu, ``Cram{\'e}r-{Rao} bound and antenna selection optimization for dual radar-communication design,'' in \emph{ICASSP 2022-2022 IEEE International Conference on Acoustics, Speech and Signal Processing (ICASSP)}.\hskip 1em plus 0.5em minus 0.4em\relax IEEE, 2022, pp. 5168--5172.

\bibitem{bengio2023gflownet}
Y.~Bengio, S.~Lahlou, T.~Deleu, E.~J. Hu, M.~Tiwari, and E.~Bengio, ``Gflownet foundations,'' \emph{Journal of Machine Learning Research}, vol.~24, no. 210, pp. 1--55, 2023.

\bibitem{bengio2021flow}
E.~Bengio, M.~Jain, M.~Korablyov, D.~Precup, and Y.~Bengio, ``Flow network based generative models for non-iterative diverse candidate generation,'' \emph{Advances in Neural Information Processing Systems}, vol.~34, pp. 27\,381--27\,394, 2021.

\bibitem{9676708}
I.~Valiulahi, C.~Masouros, A.~Salem, and F.~Liu, ``Antenna selection for energy-efficient dual-functional radar-communication systems,'' \emph{IEEE Wireless Communications Letters}, vol.~11, no.~4, pp. 741--745, 2022.

\bibitem{boyd2004convex}
S.~P. Boyd and L.~Vandenberghe, \emph{Convex optimization}.\hskip 1em plus 0.5em minus 0.4em\relax Cambridge university press, 2004.

\bibitem{wynn1972results}
H.~P. Wynn, ``Results in the theory and construction of d-optimum experimental designs,'' \emph{Journal of the Royal Statistical Society Series B: Statistical Methodology}, vol.~34, no.~2, pp. 133--147, 1972.

\bibitem{fedorov2013theory}
V.~V. Fedorov, \emph{Theory of optimal experiments}.\hskip 1em plus 0.5em minus 0.4em\relax Elsevier, 2013.

\bibitem{calinescu2011maximizing}
G.~Calinescu, C.~Chekuri, M.~Pal, and J.~Vondr{\'a}k, ``Maximizing a monotone submodular function subject to a matroid constraint,'' \emph{SIAM Journal on Computing}, vol.~40, no.~6, pp. 1740--1766, 2011.

\bibitem{tancik2020fourier}
M.~Tancik, P.~Srinivasan, B.~Mildenhall, S.~Fridovich-Keil, N.~Raghavan, U.~Singhal, R.~Ramamoorthi, J.~Barron, and R.~Ng, ``Fourier features let networks learn high frequency functions in low dimensional domains,'' \emph{Advances in Neural Information Processing Systems}, vol.~33, pp. 7537--7547, 2020.

\bibitem{9676432}
S.~Evmorfos, K.~I. Diamantaras, and A.~P. Petropulu, ``Reinforcement learning for motion policies in mobile relaying networks,'' \emph{IEEE Transactions on Signal Processing}, vol.~70, pp. 850--861, 2022.

\bibitem{10285914}
S.~Evmorfos, Z.~Xu, and A.~Petropulu, ``Gflownets for sensor selection,'' in \emph{2023 IEEE 33rd International Workshop on Machine Learning for Signal Processing (MLSP)}, 2023, pp. 1--6.

\bibitem{mildenhall2021nerf}
B.~Mildenhall, P.~P. Srinivasan, M.~Tancik, J.~T. Barron, R.~Ramamoorthi, and R.~Ng, ``Nerf: Representing scenes as neural radiance fields for view synthesis,'' \emph{Communications of the ACM}, vol.~65, no.~1, pp. 99--106, 2021.

\bibitem{kingma2014adam}
D.~P. Kingma and J.~Ba, ``Adam: A method for stochastic optimization,'' \emph{arXiv preprint arXiv:1412.6980}, 2014.

\bibitem{bourdoux20206g}
A.~Bourdoux, A.~N. Barreto, B.~van Liempd, C.~de~Lima, D.~Dardari, D.~Belot, E.-S. Lohan, G.~Seco-Granados, H.~Sarieddeen, H.~Wymeersch \emph{et~al.}, ``6g white paper on localization and sensing,'' \emph{arXiv preprint arXiv:2006.01779}, 2020.

\bibitem{10188491}
F.~Liu, L.~Zheng, Y.~Cui, C.~Masouros, A.~P. Petropulu, H.~Griffiths, and Y.~C. Eldar, ``Seventy years of radar and communications: The road from separation to integration,'' \emph{IEEE Signal Processing Magazine}, vol.~40, no.~5, pp. 106--121, 2023.

\bibitem{9747651}
Z.~Xu, F.~Liu, and A.~Petropulu, ``Cramér-rao bound and antenna selection optimization for dual radar-communication design,'' in \emph{ICASSP 2022 - 2022 IEEE International Conference on Acoustics, Speech and Signal Processing (ICASSP)}, 2022, pp. 5168--5172.

\bibitem{10097184}
L.~Xu, S.~Sun, Y.~D. Zhang, and A.~Petropulu, ``Joint antenna selection and beamforming in integrated automotive radar sensing-communications with quantized double phase shifters,'' in \emph{ICASSP 2023 - 2023 IEEE International Conference on Acoustics, Speech and Signal Processing (ICASSP)}, 2023, pp. 1--5.

\bibitem{wei2017conjugate}
Z.~Wei, W.~Chen, C.-W. Qiu, and X.~Chen, ``Conjugate gradient method for phase retrieval based on the wirtinger derivative.'' \emph{Journal of the Optical Society of America. A, Optics, Image Science, and Vision}, vol.~34, no.~5, pp. 708--712, 2017.

\bibitem{malkin2022trajectory}
N.~Malkin, M.~Jain, E.~Bengio, C.~Sun, and Y.~Bengio, ``Trajectory balance: Improved credit assignment in gflownets,'' \emph{Advances in Neural Information Processing Systems}, vol.~35, pp. 5955--5967, 2022.

\bibitem{jain2023multi}
M.~Jain, S.~C. Raparthy, A.~Hern{\'a}ndez-Garc{\i}a, J.~Rector-Brooks, Y.~Bengio, S.~Miret, and E.~Bengio, ``Multi-objective gflownets,'' in \emph{International conference on machine learning}.\hskip 1em plus 0.5em minus 0.4em\relax PMLR, 2023, pp. 14\,631--14\,653.

\end{thebibliography}

\end{document}